\documentclass[12pt, oneside]{article}
\usepackage{authblk}
\usepackage{amssymb,amsmath,amsthm, empheq}
\usepackage{mathtools} % Bonus
\usepackage{verbatim}
\usepackage[margin=1in]{geometry}
\usepackage[utf8]{inputenc}
\usepackage{csquotes}
\usepackage[linesnumbered,ruled,lined,boxed]{algorithm2e}
\usepackage{dsfont}
\usepackage[usenames, dvipsnames]{color}
\usepackage{setspace}
\usepackage{hyperref}
\usepackage[title]{appendix}
\usepackage{comment}
\usepackage[english]{babel}
\usepackage{array}
\usepackage{float}
\usepackage{multirow}
\usepackage{graphicx}
\usepackage{caption}
\usepackage{subcaption}
\usepackage{titlesec}
\usepackage{tikz}
\usetikzlibrary{positioning}
\tikzset{>=stealth}
\title{Physics-integrated generative modeling using attentive planar normalizing flow based variational autoencoder}
\author{Sheikh Waqas Akhtar}
\affil[]{University of Central Punjab, Lahore}
\affil[]{\textit {sheikh.waqas@ucp.edu.pk}}
\date{}
\begin{document}
	
\maketitle
	
%\tableofcontents
\begin{abstract}
Physics-integrated generative modeling is a class of hybrid or grey-box modeling in which we augment the the data-driven model with the physics knowledge governing the data distribution. The use of physics knowledge allows the generative model to produce output in a controlled way, so that the output, by construction, complies with the physical laws. It imparts improved generalization ability to extrapolate beyond the training distribution as well as improved interpretability because the model is partly grounded in firm domain knowledge. In this work, we aim to improve the fidelity of reconstruction and robustness to noise in the physics integrated generative model. To this end, we use variational-autoencoder as a generative model. To improve the reconstruction results of the decoder, we propose to learn the latent posterior distribution of both the physics as well as the trainable data-driven components using planar normalizng flow. Normalizng flow based posterior distribution harnesses the inherent dynamical structure of the data distribution, hence the learned model gets closer to the true underlying data distribution. To improve the robustness of generative model against noise injected in the model, we propose a modification in the encoder part of the normalizing flow based VAE. We designed the encoder to incorporate scaled dot product attention based contextual information in the noisy latent vector which will mitigate the adverse effect of noise in the latent vector and make the model more robust. We empirically evaluated our models on human locomotion dataset \cite{10} and the results validate the efficacy of our proposed models in terms of improvement in reconstruction quality as well as robustness against noise injected in the model.
\end{abstract}

\section{Introduction}
Traditional theory-driven modeling and modern data-driven modeling approaches are usually non-overlapping but recently there has been growing interest in the merger of the two, under hybrid or grey-box modeling regime. In a broad sense this includes augmenting data-driven approach with known domain knowledge. Domain knowledge is a broad term and can refer to any information about the problem. It can be the dynamics of the physical, biological or chemical system, behavioural aspect of an agent in a game, structural/mechanical property of a robotic system or rule, policies or constraints an autonomous system must obey while operating in an environment. The domain knowledge acts as a constraint for the model, for it to adapt itself as per the set boundaries. This is particularly significant in the presence of highly flexible and non-linear data-driven models which can easily overfit if their learning is not guided by domain knowledge. Domain-knowledge augmented data-driven models hold great promise towards robust models which have improved out-of-domain generalization capabilities. This hybrid regime can also play an important role towards model explainability because the decision or outcome of the model is semantically grounded in the domain knowledge. \\
Being a relatively new frontier of generative modeling, domain or (loosely speaking) physics-integrated modeling has several challenges to tackle. An important challenge is how to integrate the physics knowledge into the model learning process. Ideally, we would want to design the hybrid model such that the physics knowledge get utilized in the best possible way and would not just become redundant information in the learning process or worse cause erratic behaviour of the model. \cite{12} \cite{13} discussed such mechansims to integrate physics knowledge. Takeishi et.al \cite{4} proposed a regularized learning framework which ensures effective use of the physics knowledge. This is important because in hybrid regime, it is possible that the optimizer may over-emphasize the output of data-driven model, diminishing or even completely nullifying the output of physics model. Their model showed improved generalization ability by extrapolating to out-of-distribution scenario. \\
Another important challenge in generative modeling is to reconstruct the high dimensional signal accurately from a low dimensional latent representation of the signal. The challenge here is to learn a latent posterior distribution from a set of limited data samples which will be able to recover the true data distribution. However, learning the true posterior latent distribution is generally intractable. We usually have to use some kind of approximation. Variational inference is an approach in which the intractable posterior distribution is approximated by a base probability distribution. Many methods have been developed which use variational inference to approximate the posterior. One such method is variational auto-encoder (VAE) which is a generative model and learns an approximate latent posterior distribution parameterized by a neural network. However the choice of the class of approximate latent posterior distribution limits the representation capacity of generative model. It is evident from a number of research efforts e.g \cite{14} that approximate latent posterior distributions that are more faithful to the underlying structure in data, perform better. For example, if the data is a time series and has a dynamical structure associated to it, we would expect that the same structure be present in latent posterior distribution as well. Latent vectors sampled from such distribution, which harnesses this structure faithfully, would perform better than latent vectors which ignore the inherent structure in the data. There have also been studies which describe the detrimental effects of limited posterior approximation. \cite{15} outlines two such problems. One is the under-estimation of variance of posterior distribution which can result in incorrect and unreliable predictions. The second is that the limited capacity of posterior distribution can result in biases in the MAP estimate of model parameters. A number of approaches have been developed to learn latent posteriors that are more faithful to underlying structure of data \cite{16}\cite{17}(see section related work as well). \\
In this work, we propose to approximate latent posterior distribution in physics integrated VAE model using normalizing flows. Normalizing flow (NF) is a method for learning a probability distribution by transforming a base probability distribution (e.g a gaussian) through a series of invertible transformation called a flow. An advantage most relevant to generative modeling is that NF admits infinitesimal flow that is asymptotically able to recover true latent posterior distribution, overcoming the limitation of approaches like mean-field approximation in which no solution is ever able to recover the true posterior. Normalizing flow based latent posteriors have been used in generative models such as VAE with great success and its state of the art variants have achieved improved reconstruction results. Our motivation to use NF based posteriors in VAE model was to test its performance in a hybrid regime, and to evaluate how well it performs, if we approximate both data-driven and physics based latent posterior distribution using normalizing flow. We called this model NF-VAE hence forth.\\
Another important challenge we addressed is the presence of noise in the model. More specifically, to evaluate how the performance of model will be affected if noise is added in the feature extraction layer of the encoder. Will the noisy latent posterior be able robustly nullify the effects of noise and reconstruct the signal with high fidelity. We propose an attention based encoder architecture of NF-VAE to mitigate the effect of noise in the encoder. The idea is that if we augment the noisy latent with an additive component which is the representative of the group of latents similar to the noisy latent vector, then this would prevent the noisy latent from being too dissimilar compared to other latents of the same ilk.

%\begin{itemize}
%	\item Latent probabilistic model augment the set of observed variables with auxiliary latent variables. They are characterized by a posterior distribution over the latent variables, one which is %generally intractable and approximated by close-form alternatives. They provide an explicit parametric specification of the joint distribution over the expanded random variable space which the %distribution of observed variables is computed by marginalizing over the latent variables. VAE is one such model.
%	\item Traditional VAE everlooks the long-range correlation among latent variables, 
%	\item 
%\end{itemize}

\section{Background}\label{backgrd}
\subsection{Variational Autoencoder}\label{vae}
Variational auto-encoder (VAE) is based on amortized variational inference to approximate probability distribution $p(x)$ from which the data originated. VAE approximates data distribution $p(x)$ by a parametric distribution $p_{\theta}(x)$ with latent variable based generative process. Latent variables are produced inside the model and are generally not observable.
\begin{equation}\label{eq-vae}
p_{\theta}(x) = \int p_{\theta}(x|z)p(z)dz
\end{equation}
We assume that the prior distribution $p(z)$ on latent variable $z$ is gaussian and posterior predictive distribution is factorized bernoulli or gaussain based on the nature of data. Per-sample parameterization is performed by a neural network called decoder. The latent posterior distribution is obtained through an amortized inference in which $p_{\theta}(z|x)$ is approximated by factorized gaussian distribution $q_{\phi}(z|x)$ and the parameters $\phi$ of per-sample posterior are inferred through a neural network called encoder. Both the encoder and decoder are trained end-to-end by a gradient based optimization algorithm which maximizes the sample estimate of lower bound on evidence (ELBO) given by \ref{eq-elbo}

\begin{equation}\label{eq-elbo}
\begin{gathered}
\frac{1}{n} \sum_{i}^{n} \log p_{\theta}(x_{i}) \ge \frac{1}{n}\sum_{i}^{n}  \mathcal{L}_{\theta,\phi}(x_{i}) = \mathcal{L}_{\theta,\phi} \\
\mathcal{L}_{\theta, \phi} = \underbrace{E_{q_{\phi}(z|x)} \log p_{\theta}(x|z)}_{A} - \underbrace{KL(q_{\phi}(z|x) \| p(z))}_{B}
\end{gathered}
\end{equation}
A and B in \ref{eq-elbo}, respectively, are the negative reconstruction cost and the regularization term which penalizes the deviation of approximate posterior from the fixed prior $p(z)$.
The gradient of \ref{eq-elbo} with respect to model parameters $\theta$ can be obtained using Monte-Carlo estimation and with respect to posterior parameters $\phi$ by stochastic backpropagation using reparameterization trick \cite{7}.

\subsection{Normalizing Flows}\label{nf}
Invertible networks, also called Normalizing flows \cite{1}\cite{2}, are class of likelihood-based generative models that approximate complex distributions by warping a known base distribution (e.g a gaussian noise) through an invertible/bijective function $G: \mathrm{R}^{D} \implies \mathrm{R}^{D}$. These methods use the change of variables theorem to compute exact changes in log-density of sample after going through the bijective transformation G. Given a random variable $\textbf{z} \sim p_{z}(\textbf{z})$ the log density of $\textbf{x} = \textbf{z}_{1} = G(\textbf{z}_{0})$ follows:
\begin{equation}\label{eq1}
\ln p(\textbf{x}) = \ln p(\textbf{z}) - \ln det J_{G}(\textbf{z})
\end{equation}

If we successively apply transformation map \ref{eq1} on variables $\mathcal{z}_{k}$ with a corresponding probability distribution $q_{k}(z)$, where $k \in 0,\ldots K$, we can construct an arbitrarily complex probability density given by \ref{eq4}:
\begin{equation}\label{eq3}
\textbf{z}_{K} = g_{K} \circ \ldots \circ g_{2} \circ g_{0} 
\end{equation}

\begin{equation}\label{eq4}
\ln q_{K}(\textbf{z}_{K})= \ln q_{0}(\textbf{z}_{0}) - \sum_{k = 1}^{K}\ln \Big| det \frac{\partial f_{k}}{\partial \textbf{z}_{k-1}} \Big|
\end{equation}
where \ref{eq3} is a shorthand notation for the composition of K transformations $g_{K}(g_{K-1}(\ldots))$. The path traversed by the random variables $\textbf{z}_{k}$ with initial distribution $q_{0}(\textbf{z}_{0})$ is called the flow and the path formed by successive disbriution $q_{k}$ is a normalizing flow.

The main complexity involved in computing \ref{eq1} is the determinant of Jacobian which scales as $LD^{3}$, where L is number of hidden layers used and D is the dimension of hidden layers. Furthermore, computing the gradient of the Jacobian determinant also scales with $\mathcal{O}(LD^{3})$ and involves computing matrix inverses that can be numerically unstable.
\subsection{Planar Normalizing Flows}
Rezende and Shakir \cite{3} proposed a normalizing flow architecture with a family of bijective transformation function G of the form:
\begin{equation}\label{eq5}
 g(z) = \textbf{z} + \textbf{u}h(\textbf{w}^{T}z + b)
\end{equation}
 where $\lambda = {\textbf{w} \in R^{D}}, \textbf{u} \in R^{D}, b \in \textrm{R}$ are free parameters and $h(.)$ is a smooth element-wise non-linearity, with derivation $h^{\prime}$. Its main advantage is the cheaper Jacobian computation which takes $O(D)$ time using the matrix determinant lemma.
 \begin{equation}\label{eq6}
 \psi(\textbf{z}) = h^{\prime}(\textbf{w}^{T}\textbf{z} + b)\textbf{w}
 \end{equation}
 \begin{equation}\label{eq7}
 \Big|det \frac{\partial{g}}{\partial \textbf{z}} \Big| = |det (\textbf{I} + \textbf{u}\psi(z)^{T}) = |1 + \textbf{u} \psi(z)|
 \end{equation}
 From \ref{eq4} we conclude that density $q_{K}(z)$ obtained by transforming an arbitrary initial density $q_{0}(\textbf{z}_{0})$ through the sequence of transformation maps $g_{k}$ of the form \ref{eq5} is implicitly given:
 \begin{equation}\label{eq-zk}
 \textbf{z}_{K} = g_{K} \circ \ldots \circ g_{2} \circ g_{0}
 \end{equation}
 \begin{equation}\label{eq8} 
 \ln q_{K}(\textbf{z}_{K})= \ln q_{0}(\textbf{z}_{0}) - \sum_{k = 1}^{K}\ln \Big| 1+ \textbf{u}_{k}^{T}\psi(\textbf{z}_{k-1}) \Big|
 \end{equation}

\subsection{Scaled dot product self-attention}\label{scaled-att}
Self-attention mechanism \cite{84} represents an input sample as an attention-weighted sum of values of other input samples. The attention weight between two input samples is a scalar which determines how much one sample is similar to the other. The input samples may be words of a sentence, sequence of image frames in a video, values of a time series. In short, attention captures how a particular input sample is related to the other samples. \par
A self attention layer takes $N$ inputs $x_{1}, x_{2}, \ldots, x{N}$ each of dimension $D \times 1$ and returns $N$ output vectors of the same size. A set of values are computed for each input:
\begin{equation}\label{att-val}
v_{m} = \beta_{v} + \Omega_{v}x_{m}
\end{equation}
where $\Omega_{v} \in R^{D \times D}$ and $\beta_{v} \in R^{D}$ are weights and biases respectively. Then the nth output $SA_{n}[x_{1}, x_{2},\ldots,x_{N}]$ is a weighted sum of all the values $v_{1}, v_{2}, \dots, v_{N}$ :
\begin{equation}\label{att-out}
SA_{n}[x_{1}, x_{2},\ldots,x_{N}] = \sum_{m=1}^{N} a[x_{m},x_{n}]v_{m}
\end{equation}
The scalar weight $a[x_{m},x_{n}]$ is the attention that nth input pays to the mth input. The N weights $a[. , x_{n}]$ are non-negative and sum to 1. To compute the attention weight, we apply two more non-linear transformations to the input.
\begin{equation}\label{att-query}
q_{n} = \beta_{q} + \Omega_{q}x_{n}
\end{equation}  
\begin{equation}\label{att-key}
k_{m} = \beta_{k} + \Omega_{k}x_{m}
\end{equation}
where $q_{n}$ and $k_{m}$ are called queries and keys respectively. We can compute the dot product between queries and keys and pass the result through a softmax function:
\begin{align}\label{att-att-comp}
a[x_{m},x_{n}] &= \mathtt{Softmax}[k_{m}^{T}q_{n}] \\
 &= \frac{exp[k_{m}^{T}q_{n}]}{\sum_{m^{'}=1}^{N} exp [k_{m^{'}}^{T}q_{n}]}
\end{align}
So, the dot product is the measure of similarity between query and keys. Overall, attention weights are non-linear function of input. This is an example of hyper-parameter, in which one network computes the weights of another. The shared parameters of attention layer to learn are ${\beta_{v}, \Omega_{v}, \beta_{q}, \Omega_{q}, \beta_{k}, \Omega_{k}}$. These parameters are independent of number of inputs $N$. \par
The dot product in attention computation can have large magnitude causing inputs with large weights to dominate. Small changes in input to softmax function will have little effect on attention weight. To avoid this dot product is scaled by square root of dimension $D_{q}$ of queries or keys (both have same dimensions). \par
Summarizing the whole process in matrix form
\begin{align}\label{att-sqk}
V[X] &= \beta_{v}\textbf{1}^{T} + \Omega_{v}X \\
Q[X] &= \beta_{q}\textbf{1}^{T} + \Omega_{q}X \\
K[X] &= \beta_{k}\textbf{1}^{T} + \Omega_{k}X \\
\end{align}
where $\textbf{1}$ is $N\times1$ vector of ones. \par
Dot product self attention is computed as
\begin{equation}\label{att-dot1}
SA[X] = V[X] \texttt{Softmax}[K[X]^{T}Q[X]]
\end{equation}
Scaled dot product attention is computed as 
\begin{equation}\label{att-dot2}
SA[X] = V[X] \texttt{Softmax}\Big[\frac{[K[X]^{T}Q[X]]}{\sqrt{D_{q}}}\Big]
\end{equation}

\section{Physics Integrated generative modeling}
We propose a physics-integrated variational auto-encoding architecture for generative modeling of a dynamical system. The encoder part encodes the input signal into a latent representation. This latent approximation of the original signal is sampled from a learned posterior distribution. The decoder decodes the latent variables to reconstruct the original signal. The fidelity of reconstruction is profoundly influenced by the family of distribution, from which the latent vector is sampled. VAE uses gaussian distribution to sample latent vector in order to allow for efficient inference. However, this results in poor reconstruction results. We propose to use an attentive normalizing flow based posterior approximation to improve the reconstruction results. Normalizing flow is a generative modeling approach and has previously been used for latent posterior approximation in VAE with promising results \cite{2}. We will refer to such normalizing flow based VAE as NF-VAE. 

We now present the architecture of NF-VAE, with a brief discussion on its components:\\

\subsection{NF-VAE}
We have used NF-VAE as a hybrid generative model to learn a dynamical system. The dynamical system is such that there is a known part. We know its dynamics or physics. The other part is unknown and we don't know it dynamics. The encoder of NF-VAE produces two sets of latent variables. One for which the dynamics is known. Since we know the dynamics, the decoder of this part will just require us to solve an IVP using some numerical integrator which will take initial solution value and physics based latent vector as input. The other set of latent vector (name auxiliary variable) will belong to the unknown dynamics. Its decoder will be a neural network, mapping auxiliary latent vector to the solution of unknown ODE part in the forward process. The final output will be the sum of both decoder outputs. 
\subsubsection{Latent Variables and Prior}
We have two types of latent variables: physics based $z_{P} \in \mathcal{Z}_{P}$ and auxiliary latent variables $z_{aux} \in \mathcal{Z}_{aux}$. $\mathcal{Z}_{P}$ and $\mathcal{Z}_{aux}$ are assumed to be in Euclidean space and prior distributions on $z_{P}$ and $z_{aux}$ are assumed to be multivariate normal.
\begin{equation}\label{eq-prior}
\begin{aligned}
p(z_{P}) &= \mathcal{N}(z_{p} | \textbf{m}_{P}, \mathbf{\Sigma}_{P}) \\
p(z_{aux}) &= \mathcal{N}(z_{aux} | \textbf{m}_{aux},\mathbf{\Sigma}_{aux})
\end{aligned}
\end{equation}
$\textbf{m}_{P}$, $\mathbf{\Sigma}_{P}$ and $\textbf{m}_{aux},\mathbf{\Sigma}_{aux}$ are obtained using feature extraction neural networks $\texttt{MLP}_{\textbf{m}_{P}}$, $\texttt{MLP}_{\mathbf{\Sigma}_{P}}$, $\texttt{MLP}_{\textbf{m}_{aux}}$ and $\texttt{MLP}_{\mathbf{\Sigma}_{aux}}$ respectively.
\subsubsection{Encoder}\label{nfvae-enc}
The encoder learns latent posterior distribution of $z_{P}$ and $z_{aux}$ as:

\begin{align}\label{eq-encoder}
q_{\psi}(z_{P}, z_{aux}|x) &= q_{\psi}(z_{aux}|x)q_{\psi}(z_{P}|x, z_{aux}) \\
\text{where} \hskip 3em  q_{\psi}(z_{aux}|x) &= \ln q_{K}(z_{K}^{aux}) = \ln q_{0}(z_{0}^{aux}) - \sum_{k=1}^{K} | 1+ u_{k,aux}^{T}\psi_{k,aux}(z_{k-1}^{aux})| \\
q_{0}(z_{0}^{aux}) &= \mathcal{N}(z_{0}^{aux}|\textbf{m}_{aux}, \mathbf{\Sigma}_{P}) \\
\textbf{m}_{aux} &= \texttt{MLP}_{\textbf{m}_{aux}}(x) \\
\mathbf{\Sigma}_{aux} &= \texttt{MLP}_{\mathbf{\Sigma}_{aux}}(x) \\
q_{\psi}(z_{P}|x,z_{aux}) &= \ln q_{K}(z_{K}^{P}) = \ln q_{0}(z_{0}^{P}) - \sum_{k=1}^{K} | 1+ u_{k,P}^{T}\psi_{k,P}(z_{k-1}^{aux})| \\
q_{0}(z_{0}^{P}) &= \mathcal{N}(z_{0}^{P}|\textbf{m}_{P}, \mathbf{\Sigma}_{P})  \\
\tilde{x} &= x + \textbf{m}_{aux} \label{eq:mixing}\\
\textbf{m}_{P} &= \texttt{MLP}_{\textbf{m}_{P}}(\tilde{x}) \\
\mathbf{\Sigma}_{P} &= \texttt{MLP}_{\mathbf{\Sigma}_{P}}(\tilde{x}) 
\end{align}

The feature extracting network produces latent prior $z_{0}$ and K normalizing flow maps from $f_{0}, \ldots, f_{K}$ of the form \ref{eq6}, which transform prior latent $z_{0}$ to $z_{K}$ using \ref{eq-zk}. Features extraction and normalizing flow network makeup the recognition network or encoder of VAE. The parameters of recognition network are trained using stochastic backpropagation \cite{5}. \newline
We also used mixing process to learn physics latent $z_{p}$ grounded in auxiliary latent $z_{aux}$ \ref{eq:mixing}. This is a concatentation of mean of $z_{aux}$ with $x$ before feeding it to feature extraction network. This is important because the decoder uses both latents to reconstructs a single output. These latents cannot be completely unrelated or disjoint. The physics based latent should be grounded in auxiliary latent so that both outputs $f_{P}$ and $f_{A}$ of the decoder are aligned with each other, being exlusive parts of one whole thing and not some unrelated outputs. 

\subsubsection{Decoder}
The decoder consists of two types of functions $f_{p}: \mathcal{Z}_{p} \rightarrow \mathcal{Y}_{p}$ and $f_{aux}: \mathcal{Z}_{aux} \rightarrow \mathcal{Y}_{aux}$. We consider functional $\mathcal{F}$ which evaluates the two functions $f_{p}$ and $f_{aux}$. $f_{p}$, represents the numerically integrated dynamics of the physics model (an ODE) whereas $f_{aux}$ represents the solution of the unknown part of the dynamical system as a neural network which learns to map $z_{aux}$ to the output. The observation $x$ is the sum of both $f_{p}$ and $f_{aux}$. It may be a sequence of images or a time series. We assume observation has an gaussian noise with known variance $\sigma_{n}^{2}$ in it, hence it is also a gaussian:
\begin{align}\label{eq-dec}
p_{\theta}(x | z_{p}, z_{aux}) &= \mathcal{N} (x | \mathcal{F}[f_{p}, f_{aux}; z_{p},z_{aux}], \sigma_{n}^{2}\mathbf{I}). \\
f_{p} &= \texttt{ODESolver}(\frac{df_{p}}{dt}; t_{0}, t_{T}, x_{0}) \\ 
\frac{df_{p}}{dt} &= \texttt{MLP}_{p-decoder}(z_{p}) \\
f_{aux} &= \texttt{MLP}_{aux-decoder}(z_{aux}) \\
\end{align}
Trainable parameters of $f_{p}$ and $f_{a}$ are denoted by $\theta$. We assume additive relation between $f_{p}$ and $f_{aux}$ such that $\mathcal{F}[f_{p}, f_{aux}; z_{p},z_{aux}] = f_{p} + f_{aux}$. The role of $f_{aux}$ is complementary to the physics model in this setup. However, it can be much more than that, for example, it can also work as a correction of numerical error of ode-solver or optimizer. It can also act as side information e.g. sequence of images or video of dynamical system in its operation.
\subsubsection{Objective Function}
Following variational principle \cite{6}, we can derive a lower bound on marginal log-likelihood. This bound is often referred to as evidence lower bound (ELBO) or negative free energy.   
\begin{gather}
\ln p_{\theta}(x) = \ln \Sigma_{\mathbf{z} \sim q_{\psi}(z_{p}, z_{aux}|x)} p_{\theta} (x|\mathbf{z})p(\mathbf{z}) \\
=\ln \Sigma_{\mathbf{z}} \frac{q_{\psi} (z_{p},z_{aux} | x)}{q_{\psi} (z_{p},z_{aux} | x)} p_{\theta}(x | z_{p},z_{aux}) p(z_{p}, z_{aux})\\ 
=\ln \Sigma_{\mathbf{z}} \frac{q_{\psi}(z_{aux}|x)q_{\psi}(z_{P}|x, z_{aux})}{q_{\psi}(z_{aux}|x)q_{\psi}(z_{P}|x, z_{aux})} p_{\theta}(x | z_{p},z_{aux}) p(z_{p}) p(z_{aux})\\ 
\begin{split}
= \ln \Sigma_{z_{p}|(z_{aux},x)} \Sigma_{z_{aux}} \frac{p(z_{p})}{q_{\psi}(z_{P}|x, z_{aux})} + \ln \Sigma_{z_{p}|(z_{aux},x)} \Sigma_{z_{aux}} \frac{p(z_{aux})}{q_{\psi}(z_{aux}|x)} \\
 + \ln \Sigma_{z_{p}|(z_{aux},x)} \Sigma_{z_{aux}} p_{\theta}(x | z_{p},z_{aux}) q_{\psi}(z_{aux}|x)q_{\psi}(z_{P}|x, z_{aux})
 \end{split}
 \\
\ge E_{z_{aux}}[-D_{KL}\{q_{\psi}(z_{P}|x, z_{aux}) \| p(z_{p})\}] -D_{KL}[q_{\psi}(z_{aux}|x) \| p(z_{aux})] +  E_{\mathbf{z}} \ln p_{\theta} (x|z_{p}, z_{aux}) \\
= \mathtt{ELBO}(\theta, \psi; \textbf{x})
\end{gather}
In the last inequality we used Jensen's inequality to obtain ELBO. ELBO provides a unified objective for optimization of the model with respect to latent variables. The third term of last equation is the reconstruction error. The first and second terms are KL-divergences between approximate latents $z_{p}$ and $z_{aux}$ respectively with their corresponding priors. The divergence terms act as regularizer and try to keep the learned posterior close to prior. We can maximize the ELBO and hence the log-likelihood by minimizing the divergence terms. 
\subsubsection{Takeiski Regularizers}\label{Takeishi}
Working with two different sets of latent variable for generative modeling comes with several challenges. For example, it is possible that the trainable part of the decoder which reconstructs the unknown dynamics using a neural network, dominates in such a way that it renders the known physics model completely useless. Maximizing ELBO does not guarantee that physics knowledge is being used in an effective manner. Another challenge is that the physics based latent produced by encoder $z_{p}$ somehow becomes meaningless such that the reconstructed solution $f_{p}$ of the physics model fluctuates around the mean pattern of data. In this situation, even if the decoder effectively uses the physics model, the optimizer still would not be able to escape the local minima. To alleviate these problems, we used two additional regularizers (namely $R_{T1}$ and $R_{T2}$) proposed by Takeshi et. al (see section 3 of \cite{4}) in the objective function. Overall, the regularized objective function to optimize is:
\begin{equation}\label{overall-obh}
\text{minimize}_{\theta, \psi} -\mathbb{E}_{p_{d}(x|X)} \mathtt{ELBO} + \alpha R_{T1} + \beta R_{T2}
\end{equation}     
where $p_{d}(x|X)$ is the empirical distribution with support on data $X := \{x_{1}, x_{2},\ldots, x_{n}\}$. $\alpha$ and $beta$ are hyperparameters to control penalization by regularizers. Their optimal values were selected on the basis of performance on validation set. 
\subsubsection{Learning}
We used Adam optimizer \cite{85} to learn the model.
\subsection{Attentive NF-VAE}  
We propose a modification in NF-VAE architecture to incorporate per-sample contextual information based on scaled dot product attention mechanism. This involves an amalgamation of latent posterior with attention weighted latent posteriors. The attention weighted posteriors serves as a contextual information about the input sample, relative to other samples of the batch. Attention weighted posterior captures the relationship of posterior of a sample with posteriors of all the other samples of the batch. Incorporation of contextual attention posterior has the effect of bringing a posterior closer to the group of other similar posteriors. This will be particularly beneficial in case if the latent posterior has noise, missing values or if it is an outlier. Thus, minimizing the adverse effect of training the decoder with a noisy or outlier latent posterior. In comparison to NF-VAE, the architecture of Attentive NF-VAE has modification in the encoder only. Rest of the modules are the same as NF-VAE.  

\subsubsection{Encoder of Attentive NF-VAE} 
After getting $z_{p}$ and $z_{aux}$ from the NF-VAE encoder as described in \ref{nfvae-enc}, we pass each latent vector to its respective scaled dot product self-attention layer $\mathtt{SA}_{p}$ and $\mathtt{SA}_{aux}$  \ref{scaled-att}. We call attention weighted outputs of attention layers as $z_{p}^{att}$ and $z_{aux}^{att}$. These attention weighted output are combined with $z_{p}$ and $z_{aux}$ as following to give us attentive latent vectors.

\begin{align}
z_{p}^{att} &= \mathtt{SA}_{p}(z_{p}) \label{att-context}\\
z_{p} &= z_{p} + z_{p}\odot z_{p}^{att} \label{att-zp}\\ 
z_{aux}^{att} &= \mathtt{SA}_{aux}(z_{aux})\\ 
z_{aux} &= z_{aux} + z_{aux}\odot z_{aux}^{att}  \label{att-za}
\end{align}
where $\odot$ represent element-wise multiplication.      
      
\section{Related Work}\label{related}  

\textbf{Harnessing structure in generative modeling} \\
There have been several studies to harness different notions of structure in the generative model, with the aim to generate data more faithful to the true data distribution and improve the reconstruction fidelity. For example, mean-field approximations of latent posterior distribution incorporates a basic form of dependency with the latent variables \cite{83}. In \cite{16}\cite{6}, posterior distribution was specified as a mixture model with continuous latent variables. Posterior distribution with continuous variables \cite{7}\cite{19}\cite{21}\cite{30} and discrete \cite{25} have also been studied. Dynamical structure in latent variables was harnessed using normalizing flows in \cite{3}\cite{22}\cite{23}\cite{24}\cite{25}\cite{26}\cite{27}\cite{28}\cite{39}\cite{30}. Graph structure \cite{31}, molecules \cite{32}\cite{33}, point cloud \cite{34}, and part-model for motion synthesis using normalizing flows \cite{35} were also studied. Normalizing flow based posterior approximation for local \cite{3}\cite{26}\cite{37}\cite{38} and global \cite{36} variables have also been developed. Normalizing flows can also be used to learn posterior distribution conditioned on side information \cite{20}.  \cite{82} used attention mechanism to build more expressive variation inference model by explicitly modeling nearby and distant interaction in the latent space. \\
Based on how physics knolwedge can be utilized in the data-driven model, we can classify hybrid modeling into two main categories: \\
\textbf{Physics-integrated hybrid modeling} \\
These are the methods in which physics information is incorporated in the model design or architecture. Such models have mostly been studied in the context of prediction and generation for various applications \cite{51}\cite{51}\cite{53}\cite{58}\cite{61}\cite{66}\cite{67}\cite{68}\cite{69}\cite{46}\cite{47}\cite{70}\cite{56}\cite{71}\cite{72}\cite{54}\cite{49}\cite{73}. Further details can be found in some excellent survey papers \cite{63}\cite{64}\cite{65}. We now mention some works closest to ours in setting and construction. Takeishi et. al \cite{4} proposed a regularized hybrid model that augmented data-driven learning in VAE with physics model. Their regularizers ensure an effective utilization of physics knowledge and prevents the output of physics model from being stuck in a local minima. Yin et.al \cite{39} also proposed a regularized hybrid model. They regularized the norm $\|f_{A}\|^2$ to control the flexibility of data-driven output $f_{A}$. \cite{40}\cite{41}, although, were data-driven models, they assumed latents follow an  ODE. \cite{42} proposed a model in which latent variables, governed by hamilton mechanics, were modeled using hamiltonian neural network \cite{11}. There have also been many studies on how to integrate both physics and data-driven output components e.g they can be combined additively \cite{43}\cite{44}\cite{45}\cite{46}\cite{47}\cite{48}\cite{49} or can be combined in some composite way \cite{50}\cite{51}\cite{52}\cite{53}\cite{54}\cite{55}\cite{56}\cite{57}. Integration can also be designed in such a way that $f_{A}$ acts as a corrector to compensate for the inaccuracy or partial availability of physics knowledge or an unmodeled phenomena \cite{58}\cite{59}\cite{60}\cite{61}\cite{62}. \\
\textbf{Physics-inspired hybrid modeling} \\
These are the methods in which the physics knowledge is used to define the objective function of the model. \cite{9}\cite{13}\cite{74}\cite{75}\cite{76}\cite{77}\cite{78}\cite{79}\cite{81}. These methods assume availability of complete physics knowledge because with only partial knowledge, the objective function will also be incomplete and hence will not perform well.  
    
\section{Experiments on Human locomotion modeling}\label{experiments}
Locomotion is the movement of body from one place to another performed through a complex interaction of neuro-muscular, skeletal and sensory system. Human locomotion mainly includes walking and running. 
Modeling of human locomotion is an important challenge to be able to quantify possible deviation of the walking pattern of a patient from the physiological profiles of healthy persons, in order to do objective quantitative assessment of walking abnormalities, to develop new rehabilitation protocols and assistive devices, to personlize them according to the needs of patient and to verify their efficacy over time. 

\subsection{Dataset}
We used a subset of dataset \cite{10} which contains kinematic, kinetic and EMG measurements of locomotion at different speeds of 50 healthy subjects (25 males, 25 females, age range: 6–72 years, body mass: 18.2–110 kg,
body height: 116.6–187.5 cm). Data consisted of sequences of stride, time normalized to 100 points as a percentage of stride duration. At each time point, we extracted 3 measurements: angles of hip, knee and ankle in sagittal plane. So, each data sample $\textbf{x}$ is a sequence $\textbf{x} := [\mathbf{\omega}_{1}, \mathbf{\omega}_{2},\ldots, \mathbf{\omega}_{100}] \in \mathbb{R}^{3 \times 100}$  where $ \mathbf{\omega}_{j} := [\omega_{hip,j} , \omega_{knee,j}, \omega_{ankle,j}]^{T}$. A batch of $N$ data samples will be tensor of dimension $[N \times M \times t]$, where $N$ is the number of samples in the batch, $M=3$ is the number of measurements at each time point and $t=100$ is the number of time points of each sample.  We used 400, 100, and 344 data samples respectively for training, validation and testing.

\subsection{Models}
We experimented with the following VAE models in table \ref{exp-archs} for generative modeling on Human locomotion dataset \cite{10}. 
\begin{table}[h]
\begin{tabular} {| p{7em} | p{30em} |}
	\hline
	Ord-VAE \cite{5}\cite{7} & Ordinary VAE in which encoder neural network produces latent vector $z_{A}$. Decoder is also a neural network which reconstructs $x$ from $z_{A}$  i.e., $\mathbb{E}\textbf{x}= f_{A}(z_{A})$. No physics knowledge is used. \\
	\hline
	Phy-VAE \cite{8} & Encoder neural network produces latent vector $z_{P}$. Decoder is a physics engine which takes $z_{p}$ and $x_{0}$ as initial conditions and generates the solution for time $t$ by solving a known ODE. $\mathbb{E}\textbf{x}= f_{P}(z_{P})$ \\
	\hline
	Ord+Phy+R VAE \cite{4} & Encoder produces two sets of latent vectors $z_{A}$ and $z_{P}$ using two different neural networks. Physics based latent $z_{P}$ is reconstructed by a physics engine which know a part of dynamical system. Auxiliary latent $z_{A}$ is reconstructed by a neural network. Both reconstructed parts are added to give the final output. $\mathbb{E}\textbf{x}= f_{P}(z_{P}) + f_{A}(z_{A})$. Learning is regularized to ensure meaningful, robust latent production. \\
	\hline
	NF+Phy+R VAE(proposed) & Encoder produces two sets of latent vectors $z_{A}$ and $z_{P}$ using two different normalizing flow based neural networks. Decoder is same as in \cite{4}. Learning is regularized.\\
	\hline
	Attentive-NF+Phy+R VAE(proposed) & Encoder produces two sets of attentive latent vectors $z_{A}$ and $z_{P}$ using two different normalizing flow based neural networks. Attentive latents use contextual information capturing similarity with other clean latents. We hypothesize that embedding such contextual information in a latent makes it robust to outliers. Decoder is the same as in \cite{4}. Learning is regularized. \\
	\hline 
 \end{tabular}
\caption{Models used in the experiments}
\label{exp-archs}
\end{table}
In Ordinary VAE and Physics VAE, we used a baseline architecture similar to methods in \cite{5}\cite{7}\cite{8}. Direct comparison is not possible since our problem setting is different. With \cite{4}, we make a direct comparison. Our problem setting is the same and we have kept the same architecture in the decoder network for a fair comparison. Our proposed method differs with \cite{4} in the encoder network.   

\subsubsection{Architecture and Training details of NF-VAE and Att-NF-VAE}
\textbf{Latent Variables:} \\ We used $\mathbf{z}_{P} \in \mathbb{R}^2$ and $\mathbf{z}_{A} \in \mathbb{R}^{15}$ as physics based and auxiliary latents respectively. \\
\textbf{Learned Prior}\\
We use two distinct MLP networks for learned prior distribution $\mathcal{N}(\mu_{P}, \Sigma_{P})$ and $\mathcal{N}(\mu_{A}, \Sigma_{A})$.\\
\textbf{Physics Prior generation}:\\

$ \underbrace{M\times t + dim\_z_{A}}_\textrm{Input layer} \rightarrow \underbrace{ 512 \rightarrow 512 \rightarrow M \times t \rightarrow 512 \rightarrow 512 \rightarrow 512 }_\textrm{feature extraction layer (Feature)}$ \\ 
\begin{tikzpicture}
\node (F) at (0,0) {$\textrm{Feature}$};
\node[right=of F] (G) {$128 \rightarrow \underbrace{2}_{\mu_{P} \in R^{2}}$};
\node[below right=of F] (H) {$128 \rightarrow \underbrace{2}_{\Sigma_{P} \in R^{2}}$};
\node[above right=of H] (I) {$\underbrace{\mathcal{N}(\mu_{P}, \Sigma_{P})}_{\textrm{Latent Generation}} \sim z_{P}^{\textrm{prior}}$};
\draw[->](F)--(G);
\draw[->](F)--(H);
\draw[->](G)--(I);
\draw[->](H)--(I);
\end{tikzpicture}

\textbf{Auxiliary Prior generation}:\\
$ \underbrace{M\times t}_\textrm{Input layer} \rightarrow \underbrace{ 512 \rightarrow 512 \rightarrow 512}_\textrm{feature extraction layer (Feature)}$ \\

\begin{tikzpicture}
\node (F) at (0,0) {$\textrm{Feature}$};
\node[right=of F] (G) {$64 \rightarrow 32 \rightarrow \underbrace{15}_{\mu_{A} \in R^{15}}$};
\node[below right=of F] (H) {$64 \rightarrow 32 \rightarrow \underbrace{15}_{\Sigma_{A} \in R^{15}}$};
\node[above right=of H] (I) {$\underbrace{\mathcal{N}(\mu_{A}, \Sigma_{A})}_{\textrm{Latent Generation}} \sim z_{A}^{\textrm{prior}}$};
\draw[->](F)--(G);
\draw[->](F)--(H);
\draw[->](G)--(I);
\draw[->](H)--(I);
\end{tikzpicture}
\\
\textbf{Encoder (NF-VAE)} \\
In normalizing flow VAE (NF-VAE) encoder, we use prior latents $z_{P}^{\textrm{prior}}$ and $z_{A}^{\textrm{prior}}$ sampled from prior distribtions and apply K inverse transformations (i.e flows) $g_{k}$ sequentially to transform them to $z_{P,K}$ and $z_{A,K}$. \\
\textbf{NF Layer (Physics)}: \\
$z_{P}^{\textrm{prior}} \in \mathbb{R}^{2} \overset {g_{P,1}} {\rightarrow} z_{P,1} \overset {g_{P,2}} {\rightarrow} z_{P,2} \dots \overset{g_{P,K}}{\rightarrow} z_{P,K}$
where $g_{P,k}$ is defined as in \ref{eq-zk} \\
Similarly. \\
\textbf{NF Layer (Auxiliary)}: \\
$z_{A}^{\textrm{prior}}  \in \mathbb{R}^{15} \overset {g_{A,1}} {\rightarrow} z_{A,1} \overset {g_{A,2}} {\rightarrow} z_{A,2} \dots \overset{g_{A,K}}{\rightarrow} z_{A,K}$
where $g_{A,k}$ is defined as in \ref{eq-zk}\\
\textbf{Encoder (Attentive NF-VAE)} \\
In Attentive NF-VAE encoder, we apply a scaled dot-product attention layer after NF-layer which gives us the attentive context vector $\ref{att-context}$. We incorporate this context vector with the latent vector $z_{P,K}$ and $z_{A,K}$ of NF layer as in \ref{att-zp} and \ref{att-za} respectively, to get latents $z_{P}$ and $z_{A}$.  \\
\textbf{Decoder}: \\
In proposed models, decoder consists of two parts, one neural network based which uses $z_{A}$ and outputs $f_{A}$ and the other physics model based decoder which uses $z_{P}$ as input and outputs $f_{P}$. Reconstructed output is $\hat{x} = f_{P} + f_{A}$.\\
\textbf{Auxiliary Decoder $f_{A}$}:\\
$z_{A} \in R^{15} \rightarrow 512 \rightarrow 512 \rightarrow f_{A} \in M \times t$ \\
\textbf{Physics Decoder $f_{P}$}:\\
The decoder includes a physics engine, which takes concatenated latent $z_{P}$ and initial value $x_{0}$ as input, generates the dynamics using a known physics model. Then, a solver numerically integrates the dynamics to give us the reconstructed output. 
We modeled $\partial f_{P}$ with a trainable Hamilton equation parameterized by a neural network \cite{11}.
\begin{equation}\label{hamilton}
\partial f_{P}\Big( [\mathbf{p}^{T} \quad \mathbf{q}^{T}]^{T}, z_{P} \Big) = \Big[ -\frac{\partial \mathcal{H}^{T}}{\partial \mathbf{q}} \quad \frac{\partial \mathcal{H}^{T}}{\partial \mathbf{p}} \Big]^{T} 
\end{equation}
where $\mathbf{p} \in \mathbb{R}^{d_{H}}$ is a generalized position, and $\mathbf{q} \in \mathbb{R}^{d_{H}}$ is a generalized momentum, and $\mathcal{H}: \mathbb{R}^{d_{H}}$ is a Hamiltonian or total energy of the system. We take $d_{H}= 1$ and model $\mathcal{H}$ with an MLP with two hidden layers of size 128.  \\
Architecture of decoder is as following:\\
$\underbrace{M \times t+dim\_z_{P}}_{\textrm{Input Layer}} \rightarrow \underbrace{\partial f_{P}}_{\textrm{Physics model}\ref{hamilton}}\rightarrow \underbrace{\int}_\textrm{ODE-Solver} \rightarrow f_{P} \in M \times t$ 
\\ 
\textbf{Training settings:}
We trained for 50 epochs with a batch size of 100. We set Adam optimizer with a learning rate= $10^{-3}$, weight decay = $10^{-6}$ and eps = $10^{-3}$. Regularization hyper-parameters $\alpha$ and $\beta$ were set to $10^{-2}$ and $10^{-1}$ respectively. For NF-VAE and Att-NF-VAE, auxiliary and physics latent priors were transformed by 12 and 5 flows respectively to get auxiliary and physics latents. 

\subsubsection{Results}
\begin{figure}[h]
	\centering
	\includegraphics[width=0.95\textwidth]{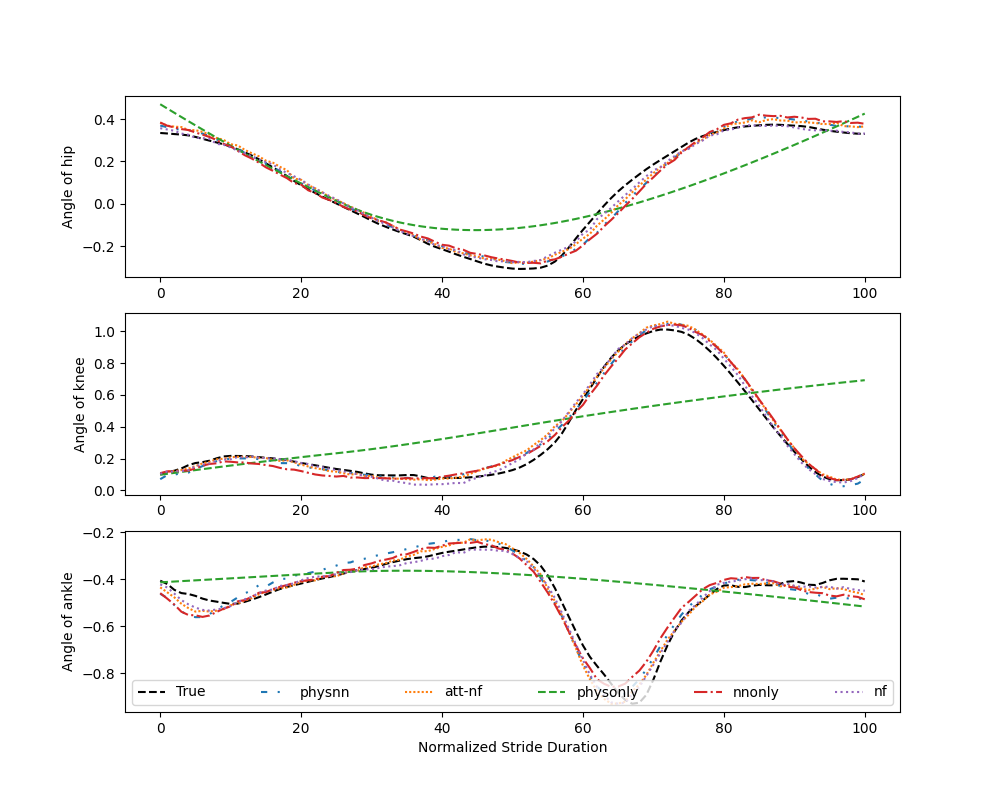}
	\caption{Reconstruction of a test sample of locomotion data.}
	\label{fig-recon}
\end{figure}

We report the mean absolute error (MAE) of the VAEs models on test data. Table \ref{table-mae} empirically demonstrate the efficacy of proposed methods over other benchmarks methods. Normalizing flow based NF-VAE with physics knowledge and regularization comes out to be the best among the lot, closely followed by Attention based NF-VAE. Physics VAE is the worst performer. Apparently, just using physics knowledge (i.e dynamcis) to reconstruct the signal is not very useful unless we integrate it with the data-driven model. Additionally, learning a rich and more expressive latent posterior distribution (NF-VAE \& Att-NF VAE) is more beneficial than assuming it to be gaussian (Ord VAE). 
	\begin{table}[H]
	\centering
	\begin{tabular}{|c|c|c|m{7em}|m{8em}|}
		\hline
		Ord VAE	& Phy VAE & Ord+Phy+R VAE & NF+Phy+R VAE(Proposed)  & Att-NF+Phy+R VAE(Proposed) \\	
		\hline
		0.2050 &  12.8470 & 0.4104 & 0.15671 & 0.2015 \\
		\hline		
	\end{tabular}
	\caption{mean absolute error (MAE) of reconstruction on human gait test data}
	\label{table-mae}	
\end{table}	
The reconstruction result of a test sample using VAEs model is shown in fig \ref{fig-recon}.
\subsubsection{Performance with noisy features}
We study the effect of noise on the top 4 VAE models of table \ref{table-mae}. We add noise by randomly zeroing some features in the last feature extraction layer. We do this by selecting first a percentage of randomly chosen samples in a batch which will be corrupted and then by zeroing a percentage of randomly chosen features in the selected samples. We trained the models for 10 epochs.

\begin{table}[H]
	\centering
	\begin{tabular}{|p{5em}|p{8em}|c|c|c|c|c|}
		\hline
		Method & \% of noisy samples & \multicolumn{5}{|c|}{\% of noisy features in each corrupt sample} \\
		\cline{3-7}
		 & & 5\% \hspace{8pt}  & 10\% \hspace{8pt} & 25\% \hspace{8pt} & 50\%  \hspace{8pt} & 75\% \\
		 \hline
		\multirow{5}{10em}{Ord VAE} & 5\% & 1.8197 & 1.8229 & 1.8259 & 1.8539 & 1.8832 \\
		\cline{2-7}
								& 10\% & 1.8204 & 1.8213 & 1.8344 & 1.8734 & 1.9125 \\
		\cline{2-7}
 							    & 25\% & 1.8186 & 1.8240 & 1.8573 & 1.9185 & 1.9377 \\
 		\cline{2-7}				
 								& 50\% & 1.8196 & 1.8317 & 1.8826 & 1.9516 & 1.9612 \\
 		\hline

	\end{tabular}
\caption{Effect of noise on MAE of Ordinary VAE}
\label{vae:noise}
\end{table}

\paragraph{Effect of noise on Ord VAE:}
Observing the trend in the mae values for various noise concentrations in table \ref{vae:noise} suggests that mae increase both by increasing the feature noise as well as the number of noisy samples. However, increase in mae by increasing feature noise is less severe when the \% of noisy samples is small.

\begin{table}[H]
	\centering
	\begin{tabular}{|p{6em}|p{8em}|c|c|c|c|c|}
		\hline
		Method & \% of noisy samples & \multicolumn{5}{|c|}{\% of noisy features in each corrupt sample} \\
		\cline{3-7}
		& & 5\% \hspace{8pt}  & 10\% \hspace{8pt} & 25\% \hspace{8pt} & 50\%  \hspace{8pt} & 75\% \\
		\hline
		\multirow{5}{6em}{Ord+Phy+R VAE} & 5\% & 1.8019 & 1.7934 & 1.7926 & 1.8163 & 1.8043 \\
		\cline{2-7}
		& 10\% & 1.8094 & 1.7920 & 1.7929 & 1.7900 & 1.8093 \\
		\cline{2-7}
		& 25\% & 1.7951 & 1.7907 & 1.8115 & 1.8298 & 1.8784 \\
		\cline{2-7}				
		& 50\% & 1.8026 & 1.7747 & 1.8560 & 1.9310 & 1.9308 \\
		\hline

	\end{tabular}
	\caption{Effect of noise on MAE of Ord+Phy+R VAE}
	\label{ord-phy-vae:noise}
\end{table} 

\paragraph{Effect of noise on Ord+Phy+R VAE:}
Observing the first two rows in table \ref{ord-phy-vae:noise} suggest that for upto  10\% of corrupt samples the mae does not increase drastically by increasing feature noise as in Ord VAE. Last two rows show gradual increase in mae on increasing feature noise, similar to Ord VAE but overall the mae values are less than Ord VAE. This establishes that Ord+Phy+R is more robust to noise than ordinary VAE. 

\begin{table}[H]
	\centering
	\begin{tabular}{|p{6em}|p{8em}|c|c|c|c|c|}
		\hline
		Method & \% of noisy samples & \multicolumn{5}{|c|}{\% of noisy features in each corrupt sample} \\
		\cline{3-7}
		& & 5\% \hspace{8pt}  & 10\% \hspace{8pt} & 25\% \hspace{8pt} & 50\%  \hspace{8pt} & 75\% \\
		\hline
		\multirow{5}{6em}{NF+Phy+R VAE} & 5\% & 1.6084 & 1.6975 & 1.6543 & 1.6885 & 1.7605 \\
		\cline{2-7}
		& 10\% & 1.7210 & 1.7695 & 1.8173 & 1.6958 & 1.7433 \\
		\cline{2-7}
		& 25\% & 1.5323 & 1.5045 & 1.6697 & 1.7263 & 1.7446 \\
		\cline{2-7}				
		& 50\% & 1.3733 & 1.5907 & 1.4378 & 1.8650 & 1.8921 \\
		\hline

	\end{tabular}
	\caption{Effect of noise on MAE of NF+Phy+R VAE}
	\label{nf-vae:noise}
\end{table}

\paragraph{Effect of noise on NF+Phy+R VAE:}
Similar to previous two models, in NF+Phy+R VAE (see table \ref{nf-vae:noise}), mae increases on increasing feature noise but the mae values are less compared to both Ord and Ord+Phy+R VAE models. We can deduce that it is more robust to both Ord and Ord+Phy+R VAE models. 

\begin{table}[H]
	\centering
	\begin{tabular}{|p{6em}|p{8em}|c|c|c|c|c|}
		\hline
		Method & \% of noisy samples & \multicolumn{5}{|c|}{\% of noisy features in each corrupt sample} \\
		\cline{3-7}
		& & 5\% \hspace{8pt}  & 10\% \hspace{8pt} & 25\% \hspace{8pt} & 50\%  \hspace{8pt} & 75\% \\
		\hline
		\multirow{5}{6em}{Att-NF+Phy+R VAE} & 5\% & 1.7679 & 1.5756 & 1.5455 & 1.8246 & 1.7387 \\
		\cline{2-7}
		& 10\% & 1.8367 & 1.8349 & 1.6837 & 1.6754 & 1.7942 \\
		\cline{2-7}
		& 25\% & 1.7348 & 1.6667 & 1.7764 & 1.8517 & 2.0694 \\
		\cline{2-7}				
		& 50\% & 1.9215 & 1.9074 & 1.7199 & 1.8315 & 1.8289 \\
		\hline

	\end{tabular}
	\caption{Effect of noise on MAE of Att-NF+Phy+R VAE}
	\label{att-nf-vae:noise}
\end{table}

\paragraph{Effect of noise on Att-NF+Phy+R VAE:}
Table \ref{att-nf-vae:noise} shows that Att-NF+Phy+R VAE has a unique behaviour on increasing feature noise (in columns) compared to other models. On increasing feature noise, mae first decreases for upto 25\% corruption in features, then increases. This can be attributed to the inclusion of attention based contextual information in the latents. Under moderate feature corruption (upto 25\%), the contextual information added to a latent would bring it closer to the uncorrupted latents, but when feature noisy is severe, contextual information becomes too corrupted itself, we see mae increasing as in other models. Overall, mae values are higher compared to NF+Phy+R VAE but less than Ord+Phy+R VAE. \\
To summarize, NF+Phy+R VAE model is the most robust to noise but Att-NF+Phy+R has a unique behaviour of decreasing mae on increasing (upto ~25\%) feature noise.

\subsubsection{Ablation Studies}
We study the effect and contribution of different factors on the performance of proposed models. \\
\textbf{Effect of latents} \\
We first study the contribution of latents $z_{P}$ and $z_{A}$, when only one is present. 
\begin{table}[H]
\centering
\begin{tabular}{|c|c|}
\hline	
NF-VAE (only $z_{A}$) & 0.1944 \\
\hline
NF-VAE (only $z_{P}$) & 13.3996 \\
\hline
Att-NF-VAE (only $z_{A}$) & 0.2347 \\
\hline
Att-NF-VAE (only $z_{P}$) & 13.9726 \\
\hline
\end{tabular}
\caption{Effect of Latents on MAE, when only one is present}	
\label{ablation:onelatent}
\end{table}
This suggests that using only data-driven approach without any physics knowledge is much better in performance than using just physics knowledge.\\
When both latents are present, where one is NF or Att-NF based, other latent is an MLP.
\begin{table}[H]
	\centering
	\begin{tabular}{|c|c|}
		\hline	
		NF-VAE $\begin{pmatrix} z_{A} \rightarrow \textrm{NF} \\ z_{P} \rightarrow \textrm{MLP} \end{pmatrix}$ & 0.1677 \\
		\hline
		NF-VAE $\begin{pmatrix} z_{A} \rightarrow \textrm{MLP} \\ z_{P} \rightarrow \textrm{NF} \end{pmatrix}$ & 0.4617 \\
		\hline
		Att-NF-VAE $\begin{pmatrix} z_{A} \rightarrow \textrm{Att-NF} \\ z_{P} \rightarrow \textrm{MLP} \end{pmatrix}$ & 0.2383 \\
		\hline
		Att-NF-VAE $\begin{pmatrix} z_{A} \rightarrow \textrm{MLP} \\ z_{P} \rightarrow \textrm{Att-NF} \end{pmatrix}$ & 0.4640 \\
		\hline
		\end{tabular}
	\caption{Effect of Latents on MAE, when one is NF/Att-NF based, other is an MLP}	
	\label{ablation:NF-MLP}
	\end{table}
When both latents are present, where one is NF based and other Att-NF based.
\begin{table}[H]
	\centering
	\begin{tabular}{|c|c|}
		\hline	
		NF-VAE $\begin{pmatrix} z_{A} \rightarrow \textrm{NF} \\ z_{P} \rightarrow \textrm{Att-NF} \end{pmatrix}$ & 0.1772 \\
		\hline
		NF-VAE $\begin{pmatrix} z_{A} \rightarrow \textrm{Att-NF} \\ z_{P} \rightarrow \textrm{NF} \end{pmatrix}$ & 0.1769 \\
		\hline
		\end{tabular}
	\caption{Effect of Latents on MAE, when one is NF based, other is a Att-NF based}	
	\label{ablation:NF-Att-NF}
	\end{table}
Analyzing the results in tables \ref{ablation:onelatent} to \ref{ablation:NF-Att-NF} and table \ref{table-mae} suggest that NF based latents perform the best.

\textbf{Effect of Regularization}\\
We discuss the effects of two Takeishi regularizers \ref{Takeishi}  on the mae performance of proposed models. Setting a hyper-parameter $=0$ nullifies the its effect. Comparing the result in table \ref{ablation:reg} with table \ref{table-mae} in which both regularizers were used, suggests that their inclusion improves the performance.  	

\begin{table}[H]
	\centering
	\begin{tabular}{|c|c|c|}
		\hline
		\multirow{3}{6em}{NF-VAE} & $\alpha = 0$ & 0.1715 \\
							 & $\beta = 0$ & 0.1766 \\
							 & $\alpha=0, \beta=0$ & 0.1775 \\
		\hline
		\multirow{3}{6em}{Att-NF-VAE} & $\alpha = 0$ & 0.2869 \\
		& $\beta = 0$ & 0.2050 \\
		& $\alpha=0, \beta=0$ & 0.2087 \\
		\hline
	\end{tabular}
	\caption{Effect of regularizers on MAE of proposed models}
	\label{ablation:reg}
	\end{table}
\section{Conclusion}
We tackled two challenges in physics-integrated generative modeling. First to improve the reconstruction performance by harnessing the dynamical structure of latent posterior distribution. Second to improve the robustness against model noise by augmenting attention based contextual information in the construction of latent posterior. Empirical evaluations validate our proposed improvements in the architecture of VAE model. In future studies, it would be interesting to investigate additional structure in the encoder (e.g latents having time-dependent dynamics). Our hypothesis is that continuous normalizing flow based posterior distribution can harness this structure. It would also be interesting to extend hybrid generative model with a complex and structured observation process, for example, partial and noisy observations \cite{18}, or observations at irregular times \cite{19}.  

\medskip

\bibliographystyle{plain}
\bibliography{reference}

\end{document}